\documentclass[letterpaper, 10 pt, conference]{ieeeconf} 

\IEEEoverridecommandlockouts                        

                            \usepackage{tikz}
\usetikzlibrary{shapes.geometric, positioning, calc}
\usepackage{graphics}
\usepackage{subcaption}
\usepackage{amsmath} 
\usepackage{amssymb}
\usepackage{cite}
\usepackage{tikz}
\usepackage{amsmath}
\usepackage{tabularx}
\usepackage{array}
\usepackage{booktabs}
\usepackage{bm}
\usepackage{algorithm}
\usepackage{algorithmic}
\PassOptionsToPackage{hyphens}{url}\usepackage{hyperref}

\usepackage{tablefootnote}
\usepackage[cal=boondox,scr=boondoxo]{mathalfa}

\DeclareMathOperator*{\argmin}{argmin}

\newcommand{\cobst}{S_{\textrm{obst}}}
\newcommand{\cfree}{S_{\textrm{free}}}
\newcommand{\cs}{s_{\textrm{s}}}
\newcommand{\cg}{s_{\textrm{g}}}
\newcommand{\st}{\textrm{s.t.}}
\definecolor{green}{RGB}{11,155,13}

\title{\LARGE \bf
Motion Memory: Leveraging Past Experiences to \\Accelerate Future Motion Planning
}

\author{Dibyendu Das${}^1$ \and Yuanjie Lu${}^1$ \and Erion Plaku${}^2$ \and Xuesu Xiao${}^1$
\thanks{${}^1$D. Das, Y. Lu and X. Xiao are affiliated with the Department of Computer Science, George Mason University, VA 22030, email: {\tt\scriptsize \{ddas6, ylu22, xiao\}@gmu.edu.}}
\thanks{${}^2$E. Plaku is affiliated with the National Science Foundation, Alexandria, VA, 22314, USA, email: {\tt\scriptsize eplaku@nsf.gov.} }
\thanks{
This work has taken place in the RobotiXX Laboratory at GMU, whose research is supported by ARO (W911NF2220242, W911NF2320004, W911NF2420027), AFCENT, Google DeepMind, Clearpath Robotics, and RTX. 
The work by E. Plaku is supported by (while serving at) the National Science Foundation. Any opinion, findings, and conclusions or recommendations expressed in this material are those of the authors and do not necessarily reflect the views of the National Science Foundation.}}

\begin{document}
\maketitle
\thispagestyle{empty}
\pagestyle{empty}

\begin{abstract}
When facing a new motion-planning problem, most motion planners solve it from scratch, e.g., via sampling and exploration or starting optimization from a straight-line path. However, most motion planners have to experience a variety of planning problems throughout their lifetimes, which are yet to be leveraged for future planning. In this paper, we present a simple but efficient method called \emph{Motion Memory}, which allows different motion planners to accelerate future planning using past experiences. Treating existing motion planners as either a closed or open box, we present a variety of ways that Motion Memory can contribute to reduce the planning time when facing a new planning problem. We provide extensive experiment results with three different motion planners on three classes of planning problems with over 30,000 problem instances and show that planning speed can be significantly reduced by up to 89\% with the proposed Motion Memory technique and with increasing past planning experiences.

\end{abstract}

\section{INTRODUCTION}
\label{sec::introduction}
Motion planning refers to the computational process of determining a sequence of control inputs and actions to move a robot from a given start  state to a desired goal location while avoiding obstacles and observing system and environment constraints. 
Motion planners are essential components for almost all robotic applications~\cite{canny1988complexity}, such as autonomous navigation~\cite{kuwata2009real},~\cite{fox1997dynamic} and manipulation~\cite{murray2016robot},~\cite{volpe2003rover}. Therefore, quick, efficient, and optimal collision-free motion planning is of paramount value to the entire robotics community. 

Decades of research into motion planning have made significant progress and are able to find motion-planning solutions for different robot platforms, e.g., using Probabilistic Roadmaps (PRM)~\cite{kavraki1996probabilistic}, Expansive Spaces algorithm~\cite{hsu1997path}, and Rapidly-Exploring Random Trees (RRT)~\cite{lavalle1998rapidly, karaman2011sampling} to move mobile robots or manipulator arms. Nevertheless, these planners still face challenges in complex real-world settings when real-time planning is required to assure fast and reliable motion execution. Conventional motion planners need to plan from scratch every time they encounter a new environment. This situation remains true even when robots repeatedly face similar environments, where prior experiences could be beneficial. Such repetitive planning introduces unnecessary planning time and therefore limits the robot performance in real-world environments where fast planning time can benefit the downstream tasks, such as quickly moving through highly constrained obstacle spaces. 

On the other hand, advances in machine learning have demonstrated that robots are capable of learning emergent behaviors in a data-driven manner without depending on heavily engineered attributes and heuristics. One particular benefit of learning methods is the potential to continually improve with increasing real deployment experiences~\cite{xiao2022motion}, a capability that the classical motion planners lack. 

\begin{figure}[t]
    \centering 
    \includegraphics[width=0.8\columnwidth]{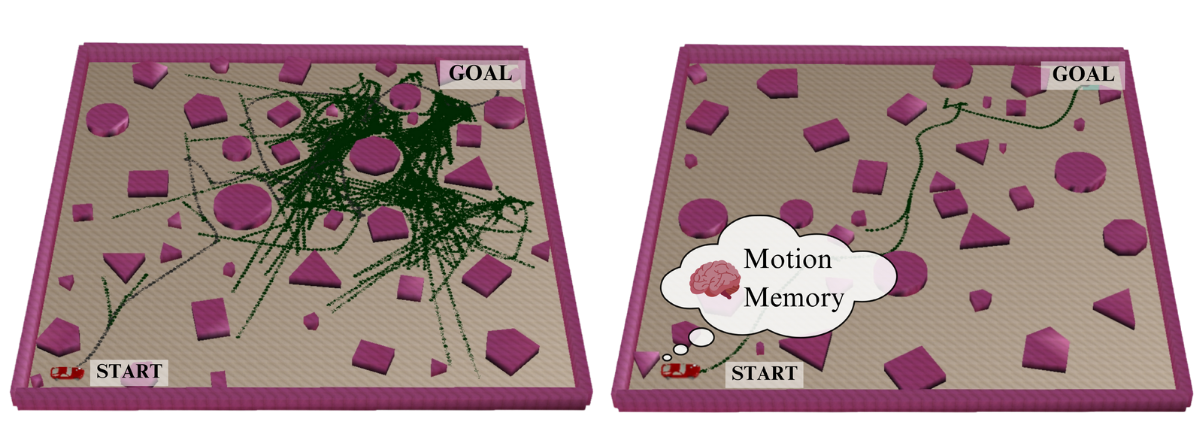} 
    \caption{Traditional motion planners require significant amount of effort to plan from scratch (left), such as large amount of samples or iterations (illustrated in green); Motion Memory utilizes past planning experiences to accelerate future planning when facing new planing problems (right). }
    \label{fig:intro}
\end{figure}

Considering the limitations of classical motion planners and the potential of learning from experiences, we present Motion Memory, a new paradigm based on past planning experiences to guide traditional motion-planning methods when facing new planning problems in order to reduce computational overhead and therefore improve planning efficiency, as robots gather more and more deployment experiences in the real world. Leveraging machine learning, Motion Memory includes an experience augmentation technique and a representation learning method that enable robots to reflect on prior planning experiences for efficient future planning as shown in Fig.~\ref{fig:intro}. 
To be specific, the experience augmentation strategy automatically generates new planning problems, for which past motion plans are (or are not) the solutions, and thus provides Motion Memory with an extensive corpus of training data to generalize to future planning problems.
Motion Memory also utilizes representation learning to enable autonomous robots to learn from augmented previous planning experiences so that motion planners can identify, store, memorize, and retrieve past planning experiences to facilitate motion planning in unseen future environments. 
We present different ways to integrate Motion Memory with three existing motion planners in three different categories of environments both in a closed and open box manner to showcase the wide applicability and generalizability of the technique. 
Our experiments demonstrate that Motion Memory significantly reduces the motion-planning time in future unseen environments by up to 89\%  with increasing deployment experiences.
\section{RELATED WORK}
\label{sec::related_work}
We review related work in classical motion planning and motion planning with machine learning. 
\subsection{Classical Motion Planning}
Among classical motion planners, sampling-based approaches have shown to be effective in solving challenging problems in complex, unstructured environments \cite{book:MP,book:LaValle}. To account for dynamics~\cite{xiao2021learning, karnan2022vi, atreya2022high, datar2023learning}, sampling-based motion planners often expand a motion tree whose branches correspond to collision-free and dynamically-feasible trajectories. RRT\cite{RRT,RRTRecent1} and its variants~\cite{RRTguided,RRTreach,RRTtrans2} rely on nearest neighbors to expand the motion tree toward random samples. EST~\cite{EST} relies on probability distributions to push the tree toward less-explored areas. KPIECE~\cite{KPIECE} leverages a grid decomposition and interior-exterior cells. GUST~\cite{GUST} introduces a discrete layer based on a roadmap abstraction and relies on discrete search to expand the motion tree along shortest paths in the roadmap. The work in \cite{Follow} further improves the motion-tree expansion to more aggressively follow the roadmap paths, while also increasing the clearance from the obstacles.   When presented with a new planning problem, these approaches, however, have to plan from scratch as they do not learn from prior experiences. This is precisely what our Motion Memory framework addresses, enabling motion planners to leverage prior solutions to similar planning problems.

\subsection{Learning Assisted Motion Planning}
Throughout the years, the learning and planning communities have investigated a variety of strategies for implementing the concept of machine learning for efficient motion planning~\cite{mcmahon2022survey}. One intuitive approach is to apply machine learning to generate high-quality valid samples in critical regions more efficiently. Some methods produce valid samples in relevant areas by learning the representation of the configuration space~\cite{aoude2013probabilistically,burns2005sampling,kingston2019exploring,sutanto2021learning}, whereas other methods train models to bias sampling over the critical regions of predicted trajectories~\cite{baldwin2010non,zucker2008adaptive,ichter2018learning}. Another practical strategy is to minimize the number of collisions~\cite{hauser2015lazy, mandalika2019generalized, bialkowski2016efficient} by learning Gaussian mixture models~\cite{huh2016learning}, kernel perceptron~\cite{das2020learning}, and graph neural network~\cite{yu2021reducing} to replace the conventional collision checker or using a learned model to determine the order of testing the nodes to identify valid paths~\cite{pan2013faster,bhardwaj2021leveraging,hou2020posterior}. 

Instead of learning a specific planning operation, other learning-based pipelines learn from prior experience to solve motion planning problems in an end-to-end manner. One strategy is to use similarity function to determine the relevant information and retrieve it from a database in the form of paths~\cite{berenson2012robot, pairet2021path,coleman2015experience} or sampling distributions~\cite{chamzas2019using, finney2007predicting}. Another approach is to train a deep neural network for efficient motion planning by using a database of past solved problems. These methods take as input point-cloud representations of a workspace and learn to encode this point-cloud into a latent space~\cite{qureshi2019motion, johnson2020dynamically, strudel2021learning, qureshi2020neural}. Chamzas et al.~\cite{chamzas2021learning} and Lien et al.~\cite{lien2009planning} construct a similarity function only over the workspace to extract suitable local representations of planning problems~\cite{chamzas2022learning}. 

Informed by the aforementioned works, this paper proposes to hallucinate planning problems by experience augmentation, construct a memory of past planning experiences by classifying environments as similar or dissimilar using representation learning, and retrieve relevant planning experience to assist planners in future planning.
Compared to all these specific techniques developed for a specific motion planner in an ad hoc manner, our Motion Memory is designed to be a universal technique that is applicable to any motion planner and allows them to improve planning efficiency with increasing planning experiences in a principled manner.

\section{Motion Memory}
\label{sec::approach}

The key inspiration behind Motion Memory is that most motion planners will encounter many different planning problems and produce different planning solutions throughout their lifetimes. Motion Memory aims to reflect on past planning experiences by generating a set of planning problems that will make the past planning solutions feasible/non-feasible. Then, utilizing representation learning, Motion Memory learns a latent space which contains efficient but also expressive information regarding the feasibility or optimality of the existing problem solutions with respect to any new planning problems in the future. 

 
\begin{figure*}[t]
\centering
\includegraphics[width=0.7\textwidth]{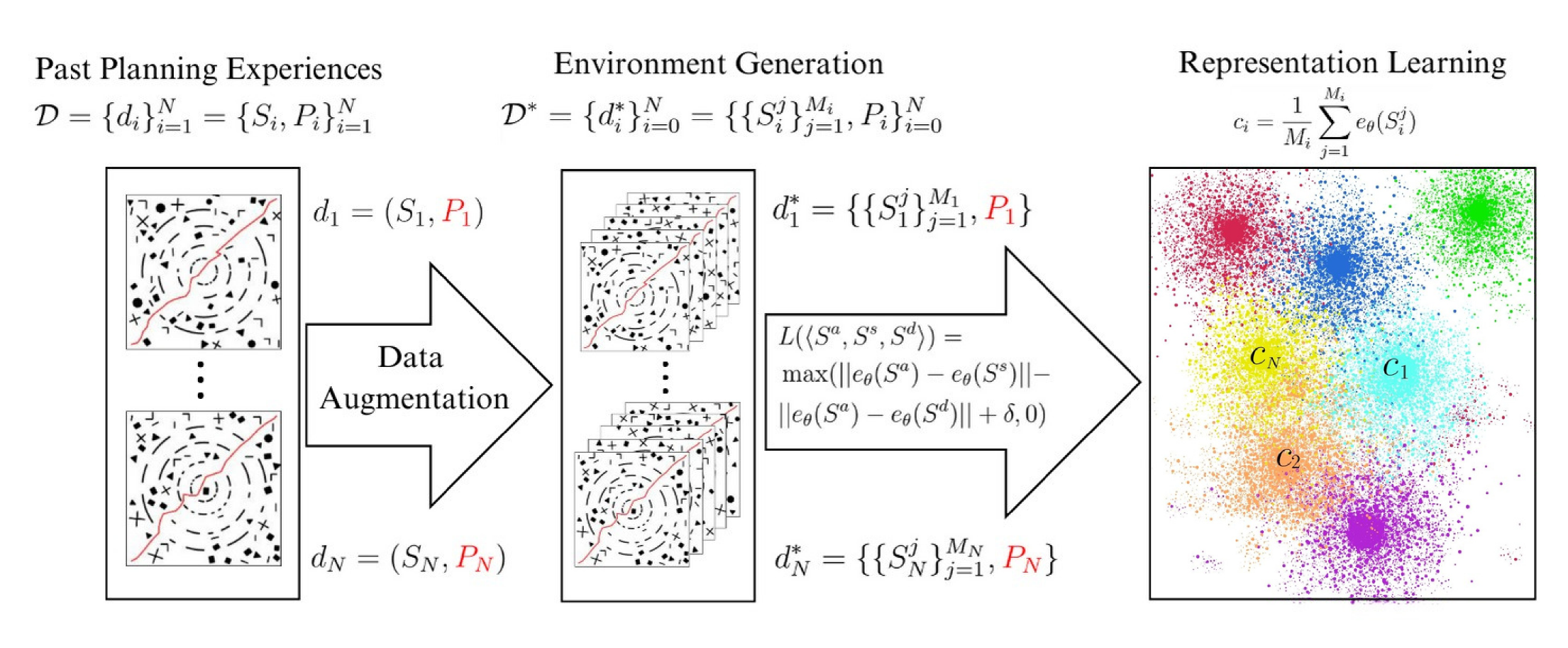}
\caption{Motion Memory: Using Past Planning Experiences, Environment Generation, and Representation Learning to Accelerate Future Planning.}
\label{fig:Approach}
\end{figure*}

\subsection{Problem Formulation}
We formulate our motion planning problem in the state space (S-space)~\cite{latombe2012robot}, which represents the universe of all possible robot states. 
For the scope of this paper, we use a 2D ground mobile robot as an example, but the general paradigm of Motion Memory has the potential to scale up to high-dimensional workspaces. 
A particular environment's S-space can be decomposed as $ S = \cfree \cup \cobst $, where $\cfree \in \mathcal{S}_{\textrm{free}}$ is the set of reachable states and $\cobst \in \mathcal{S}_{\textrm{obst}}$ is the unreachable set due to obstacles, nonholonomic constraints, velocity bounds, etc. 
The robot state $s \in S$ is defined as in collision with obstacles when $s \in \cobst$ and collision free otherwise ($s \in \cfree$). 
We denote the robot action to be $u\in\mathcal{U}$, e.g., commanded linear and angular velocity $\left( v, \omega \right)$, and define a motion plan $P \in \mathcal{P}$ as a sequence of such actions and resulted states $P=\{u_i, s_i~|~1\leq i \leq t\}$. 
$\mathcal{P}$ is the space of all motion plans. 
Notice that some motion planners may produce the resulted states only, i.e., $P=\{s_i~|~1\leq i \leq t\}$ and leave the action generation to a low-level controller. 
With the aforementioned notation, any motion planner can be defined as a function $f(\cdot)$ that can be used to produce motion plans,
\begin{equation}
\begin{gathered}
    P=f(S~|~\cs, \cg), \\
    \st \quad S = \cfree \cup \cobst, \quad \cs, \cg, s_i \in \cfree, \forall s_i \in P,  
\end{gathered}
\label{eqn::mp}
\end{equation}
that result in the robot moving from the robot's start state $\cs$ to a specified goal state $\cg$ without intersecting $\cobst$, while observing robot motion constraints and optimizing a particular cost function (e.g. distance, clearance, energy, and combinations thereof). 

A motion planner $f(\cdot)$ that has been used to solve different motion planning problems in the past will have available a dataset $\mathcal{D}=\{d_i\}_{i=1}^N=\{S_i, P_i\}_{i=1}^N$, where $N$ is the amount of currently available S-space (planning problem) and motion plan (planning solution) pairs. 
Here, we assume that the start and goal states are a constant distance away from each other and remain the same throughout all planning problems, e.g., by aligning them through rotation and translation of the S-space, a realistic assumption in many real-world motion planning problems where dynamics is primarily considered in the local planner, which is guided by local goals a constant distance away from the robot computed by a global planner. For example, $\cs$ can be always expressed as the origin of the S-space, and $\cg$ as $(50, 50)$. The goal of Motion Memory is to utilize the existing $\mathcal{D}$ from past planning experiences to accelerate future planning as shown in Fig.~\ref{fig:Approach}.
\vspace{-5pt}
\subsection{Augmenting Past Experiences with Hallucination}
Although $\mathcal{D}$ may contain a variety of planning problem and solution pairs which have the potential to cover a new planning problem faced in the future, the chance of encountering exactly the same problem is still very low, considering the changing real world,  sensory noise, and perceptual imperfectness. Therefore, Motion Memory bootstraps on the existing dataset $\mathcal{D}$ and augments it with hallucinated planning problems. 

In contrast to the motion planning problem of finding the optimal motion plan $P$ given a S-space $S$ in Eqn.~(\ref{eqn::mp}), the hallucination problem is defined as its inverse problem~\cite{xiao2021toward, xiao2021agile, wang2021agile}, i.e., given a motion plan $P$, what are the S-spaces $S$ that assure this motion plan is optimal: 
\begin{equation}
\begin{gathered}
    \{S_i\}_{i=1}^\infty = f^{\dagger}(P~|~\cs, \cg),\\
    \st \quad S = \cfree \cup \cobst, \quad \cs, \cg, s_i \in \cfree, \forall s_i \in P. 
\end{gathered}
\label{eqn::hallucination}
\end{equation}
Notice that $f^{\dagger}(\cdot)$ is not strictly the inverse function of $f(\cdot)$, because $f(\cdot)$ is non-injective, i.e., different S-spaces can map to the same optimal plan. Therefore, instead of one S-space $S$, we need to find the set of all S-spaces $\{S_i\}_{i=1}^\infty$, where the motion plan $P$ is optimal. In most cases, such a set is an infinity set, except for finitely discretized $S$. 

Compared to the forward motion planning problem, which needs to run in real time onboard computation-limited robot platforms, the inverse hallucination problem in Eqn.~(\ref{eqn::hallucination}) is much easier to solve. Researchers have proposed different ways to approximate the infinity set $\{S_i\}_{i=1}^\infty$ using representative obstacle states, e.g., the maximal~\cite{xiao2021toward}, a minimal~\cite{xiao2021agile}, or a learned distribution of~\cite{wang2021agile} obstacle set in the S-space. With these hallucination techniques, each member of the  original pairwise motion planning problem-solution dataset can be augmented from $d_i= \{S_i, P_i\}$ to  $d_i^*= \{\{S_i^j\}_{j=1}^{M_i}, P_i\}$, where $M_i$ is the number of total generated planning problems for solution $i$. Thus, the original dataset $\mathcal{D}$ is augmented to 
\begin{equation}
    \mathcal{D}^* = \{d_i^*\}_{i=0}^N = \{ \{S_i^j\}_{j=1}^{M_i}, P_i\}_{i=0}^N, 
    \nonumber
\end{equation}
in which each past planning solution $P_i$ is no longer only paired with one original planning problem $S_i$, but a set of $M_i$ planning problems $\{S_i^j\}_{j=1}^{M_i}$, for which $P_i$ is optimal. 
While $P_1, P_2, ..., P_N$ indicate $N$ existing motion plans, $P_0$ corresponds to an empty set: 
It is also possible to produce a set of motion planning problems in which no existing motion planning solutions are feasible or optimal, e.g., by adding obstacles to intersect with at least one robot state in every existing motion plan,  which will yield $d_0^* = \{\{S_0^j\}_{j=1}^{M_0}, P_0\}$, which is also included in $\mathcal{D}^*$. 

\subsection{Representation Learning}
The augmented dataset $\mathcal{D}^*$ contains more planning problems for each past planning solution, where the solution is feasible or optimal. Such an augmented dataset is used to learn an efficient latent representation space, which contains critical information regarding the feasibility or optimality of each existing motion plan solution with respect to any motion planning problem. Specifically, we adopt a triplet loss to enforce solution invariance in the learned latent space so that all environments where solution $P_i$ is feasible or optimal stay close to each other in the learned embedding space. We generate triplet training data of anchor, similar, and dissimilar planning problems $\langle S^a, S^s, S^d \rangle$ by sampling $S^a$ and $S^s$ from the same set of planning problems where solution $P_i$ is feasible or optimal, i.e., $\{S_i^j\}_{j=1}^{M_i}$, and sample $S^d$ from other problem sets and assure $P_i$ is not feasible or optimal for $S^d$. The planning problem (S-space) encoder $e_\theta(\cdot)$ is then trained to minimize a triplet loss, 
\begin{equation}
    \begin{split}
        &L(\langle S^a, S^s, S^d \rangle) = \\
&\max(||e_\theta(S^a) - e_\theta(S^s)|| - ||e_\theta(S^a) - e_\theta(S^d)|| + \delta, 0), 
    \end{split}
    \label{eqn::triplet}
    \nonumber
\end{equation}
with $\theta$ as the learnable parameters. The representation space will contain data points from $N$ different latent clusters corresponding to the $N$ existing motion plans $\{P_i\}_{i=0}^N$, including a cluster for S-spaces where none of the $N$ plans are feasible or optimal. The cluster centroids are computed as 
\begin{equation}
    c_i = \frac{1}{M_i}\sum_{j=1}^{M_i} e_\theta(S_i^j).
    \nonumber
\end{equation}

\subsection{Accelerating Future Planning}
Using representation learning, the planning problem encoder is able to project any new motion planning problem $S_{N+1}$ into the latent space, i.e., $l_{N+1} = e_\theta(S_{N+1})$. The existing motion plan associated with the closest cluster centroid $P_{i^*}$ is likely the closest to the actual motion planning solution of the problem $S_{n+1}$:
\begin{equation}
    i^* = \argmin_{i} ||e_\theta(S_{n+1}) - c_i||. 
    \nonumber
\end{equation}
For existing motion planners, a motion plan $P_{i^*}$ potentially very close to the actual solution can be used to accelerate planning in different ways. For example, for sampling-based motion planners, the states in $P_{i^*}$ can be used to bias sampling. 
For optimization-based approaches, both the actions and states can be used as an initial guess, highlighting the potential of Motion Memory to serve as an effective seed~\cite{lembono2020memory}, although this specific application is not demonstrated in our experiments.
For motion planners treated as a closed-box or an open-box, $P_{i^*}$ can be collision checked and be used as the final solution if no collisions are detected, or utilize the closed-box planner to only fix the plan segments which are in collision. In cases where $P_0$ is the closest in the representation space, the motion planner can start planning from scratch.

\section{EXPERIMENTS}
\label{sec::experiments}

\begin{figure}[t]
    \centering 
    \includegraphics[width=0.9\columnwidth]{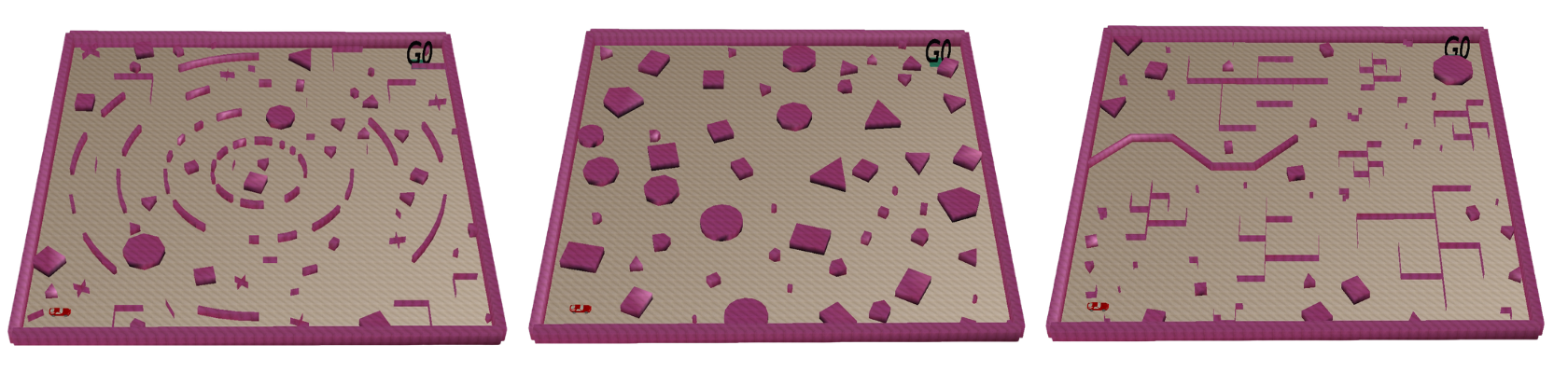} 
    \caption{Different Planning Problem Classes : Curves (Left), Random (Middle), Trap (Right)}
    \label{fig::env}
\end{figure}

\begin{figure*}[t]
\centering
\includegraphics[width=0.8\textwidth]{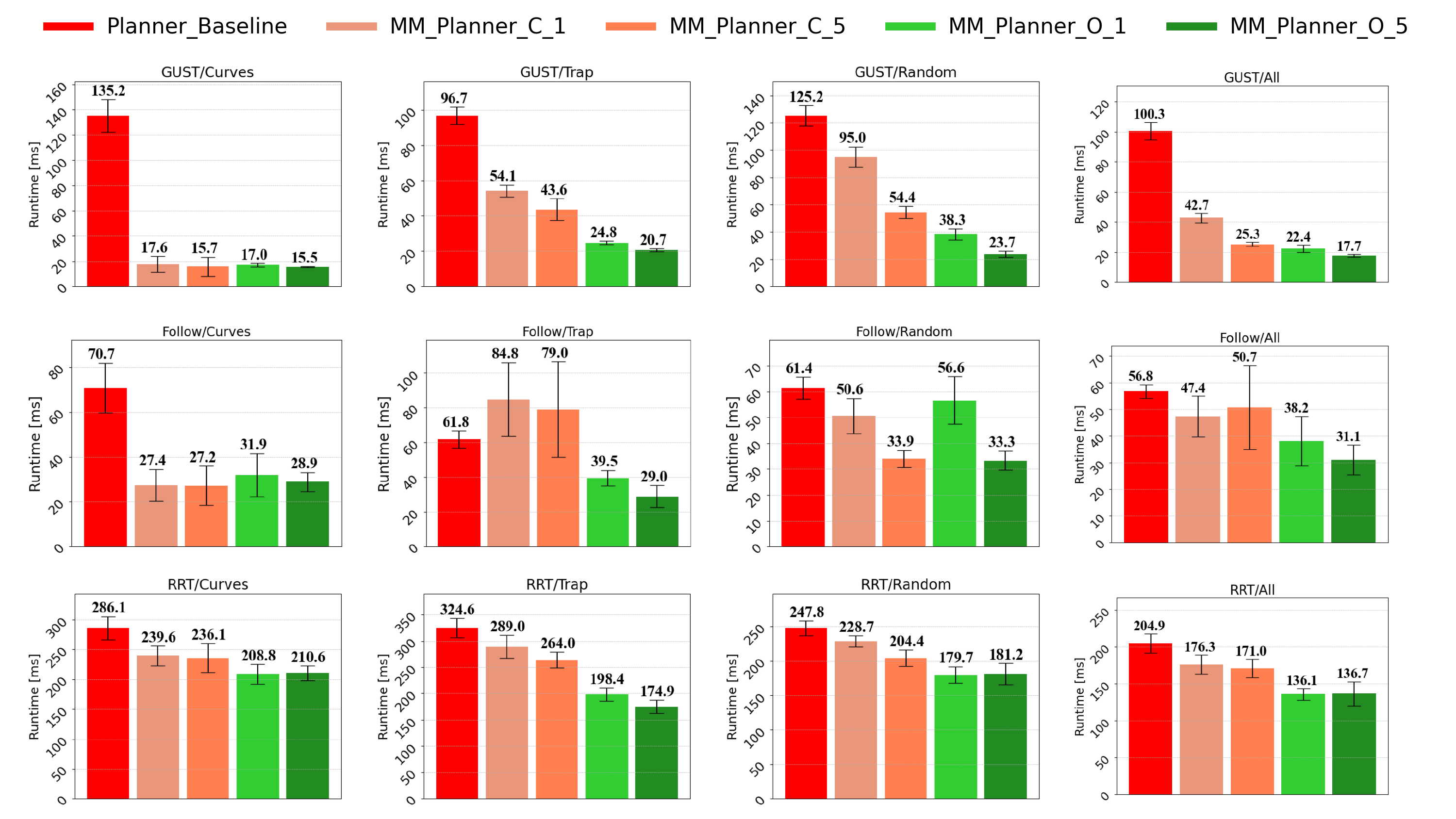}
\caption{Different Motion Memory configurations improve three planners in three classes of motion planning problems.}
\label{fig::12_figures}
\end{figure*}

We present extensive experimental results by integrating our Motion Memory technique with three different motion planners in both a closed-box and an open-box manner to solve three different classes of motion-planning problems. We present detailed results on the three planners' performances with or without Motion Memory for each problem class and for all three classes combined into one. Furthermore, we present evidence demonstrating Motion Memory allows motion planners' planning efficiency to improve with increasing past planning experiences. Finally, we conduct an ablation study to show that the improved planning efficiency is not only due to more past planning experiences, rather the Motion Memory technique itself is an indispensable part. 

\subsection{Experimental Setup}
Considering Motion Memory is designed to be a universal paradigm that is agnostic to any underlying planner, we integrate Motion Memory with three different motion planners, in both a closed-box and an open-box manner to show its general applicability (denoted as ``C'' and ``O'' in all experiment results). In the closed-box integration, we do not assume any access to or knowledge about the underlying planner, and only interface Motion Memory with it using the Motion Memory output as a potential solution after taking a new planning problem as input. 
The open-box integration assumes access to and knowledge about the underlying planner and depends on which components can be benefited from the Motion Memory output, e.g., sampling distribution or initial optimization guess. We describe the three underlying planners with their closed-box and open-box Motion Memory integration as follows.  

\subsubsection{Baseline Planners}

We experiment with different sampling-based motion planners. We first use RRT~\cite{RRT,RRTRecent1}, one of the most popular methods. RRT uses random sampling and nearest neighbors to guide the motion-tree expansion. We also use GUST~\cite{GUST}, which was specifically designed for motion-planning problems for vehicles with dynamics. GUST introduces a discrete layer obtained by building a roadmap to guide the motion-tree expansion. 
The roadmap is then used to induce a partition of the motion-tree into groups based on their nearest roadmap node. 
During the motion-tree expansion, priority is given to those groups associated with short paths to the goal. Our third planner, which we refer to as Follow~\cite{Follow}, further improves GUST by more aggressively following the roadmap paths, and dynamically adjusting the weights based on the progress made during the motion-tree expansion.

\paragraph{Closed-box Versions of the Motion Planners}
We first use these planners as black boxes in conjunction with the motion-memory framework. Specifically, let $\zeta_1, \ldots, \zeta_k$ be the top $k$-predictions obtained by the motion-memory framework for a new motion-planning problem. Each of these predictions corresponds to a dynamically-feasible trajectory (retrieved from the database). The black-box version, referred to as $\text{MP}_\text{ClosedBox}$, first checks these trajectories $\zeta_1, \ldots, \zeta_k$ in order for collisions. If some $\zeta_i$ is not in collision, then $\text{MP}_\text{ClosedBox}$ returns $\zeta_i$ as the solution. Otherwise, $\text{MP}_\text{ClosedBox}$ runs $\text{MP}$ and returns the solution (if any) found by $\text{MP}$.

\paragraph{Open-box Versions of the Motion Planners}
We can better leverage the predictions made by the motion-memory framework by looking inside the motion planners to better leverage the predictions. We refer to these versions as $\text{MP}_\text{OpenBox}$.
$\text{MP}_\text{OpenBox}$, as the closed-box version, starts by checking if any of the predicted trajectories $\zeta_1, \ldots, \zeta_k$ is not in collision. If all of them are in collision, then $\text{MP}_\text{OpenBox}$ runs a modified version of $\text{MP}$, as described below.

$\text{RRT}_\text{OpenBox}$ is obtained by changing the sampling distribution from which the target is drawn. In the original RRT, the target is sampled with probability $b_\text{goal}$ (a predetermined probability, usually $b_\text{goal}=0.1$) from an area near the goal, and with probability $1-b_\text{goal}$ from the entire state space. To leverage the predictions, we bias the sampling to be along the predicted trajectories $\zeta_1, \ldots, \zeta_k$. In particular, let $b_1, \ldots, b_k, b_\text{goal}, b_\text{other}$ such that $b_1 + \ldots + b_k + b_\text{goal} + b_\text{other} = 1$. Generally, $b_1 > \ldots > b_k > b_\text{goal} > b_\text{other}$. We sample the target with probability $b_i$ along $\zeta_i$, with probability $b_\text{goal}$ from near the goal, and with probability $b_\text{other}$ from the entire state space. To generate a target along $\zeta_i$, we first select an intermediate state in $\zeta_i$ and then sample around it. In this way, $\text{RRT}_\text{OpenBox}$ biases the exploration toward the predicted trajectories $\zeta_1, \ldots, \zeta_k$.

$\text{GUST}_\text{OpenBox}$ differs from $\text{GUST}$ only in the roadmap construction. Specifically, while $\text{GUST}$ generates a roadmap node by sampling from the entire space, $\text{GUST}_\text{OpenBox}$ biases the sampling along  $\zeta_1, \ldots, \zeta_k$. Specifically, to generate a roadmap node, $\text{GUST}_\text{OpenBox}$ selects $\zeta_i$ with probability $b_i$, selects a state $s$ uniformly at random from $\zeta_i$, and generates a roadmap node by sampling near $s$. This results in a roadmap that captures the connectivity along the predicted trajectories $\zeta_1, \ldots, \zeta_k$. Note that the roadmap is collision free, even if parts of $\zeta_1, \ldots, \zeta_k$ are in collision (since samples that result in collision are discarded during the roadmap construction). This biased roadmap construction allows for a more efficient expansion of the motion tree along the predicted trajectories.

$\text{Follow}_\text{OpenBox}$ differs from $\text{Follow}$ only in the roadmap construction. In fact, $\text{Follow}_\text{OpenBox}$ uses the same procedure as $\text{GUST}_\text{OpenBox}$ for constructing the roadmap.  $\text{Follow}_\text{OpenBox}$ then seeks to closely follow the shortest roadmap path from the start to the goal by aggressively expanding the motion tree to follow the nodes in the shortest path in succession.

\subsubsection{Planning Problem Classes}
Fig.~\ref{fig::env} shows the three different classes of planning problems used in our experiments, referred to as ``Curves," ``Random," and ``Trap." These problem classes are parametrized, and the obstacles are procedurally generated. Placement (location and orientation) and size of the obstacles are drawn from a probability distribution so that numerous instances can be generated for each problem class. An instance corresponds to a specific placement of the obstacles, as shown in Fig.~\ref{fig::env}. For example, for the ``Curves" problem class we can vary the number of curves, number of segments per curve, separation among segments, and so on. For the ``Random" problem class, we can vary the density of the obstacles as well as their shape and placement. For the ``Trap" problem class we can perturb the placement of the major obstacles and also of the random obstacles spread throughout the environment.

\subsubsection{Motion Memory Implementation}
For each of the three motion planning problem classes, we assume our Motion Memory has access to 100 planning problems and their corresponding motion planning solutions as its past planning experiences, i.e., $\mathcal{D}=\{d_i\}_{i=1}^{100}=\{S_i, P_i\}_{i=1}^{100}$. We augment each data point in $\mathcal{D}$ with 999 more planning problems and generate an augmented dataset $\mathcal{D}^* = \{d_i^*\}_{i=1}^{100} = \{ \{S_i^j\}_{j=1}^{1000}, P_i\}_{i=1}^{100}$, a dataset of 100,000 problems per class with 100 solutions, by slightly rearranging obstacles close to the motion plan and randomly shuffling obstacles in other places. For simplicity, we omit the set of problems where no existing motion plan is a solution, i.e., $d_0^*$. To train the planning problem encoder $e_\theta(\cdot)$, we use a Convolutional Neural Network 
which takes motion planning problems as input and outputs a 30-dimensional latent space. Specifically, the input to our CNN is a discretized version of the workspace, represented as a 2D grid where each cell encodes the presence or absence of an obstacle, and the network architecture comprises sequential layers of convolution and pooling, and the final feature map is processed through fully connected layers to produce the 30-dimensional embedding. For testing, we generate another 10,000 unseen planning problems per class for the motion planners to solve. 

We present experiments comparing closed-box and open-box integrations with both top-one and top-five predictions for new problems. While top-five predictions provide more information, they also incur additional computational costs due to the need to process multiple predicted solutions.


\subsection{Improvement for Different Planners and Problems}
In Fig.~\ref{fig::12_figures}, we show individual planning performance of the three underlying planners (GUST, Follow, and RRT in row 1, 2, and 3 respectively) in the three classes of motion planning problems (Curves, Random, and Trap in column 1, 2, and 3 respectively). The fourth column is the results of a large Motion Memory which does not distinguish among different motion planning problem classes, with 300,000 training and 30,000 testing data points. In each figure, we show the average planning time of default baseline planner without Motion Memory (\texttt{Planner\_Baseline}), the baseline planner assisted by Motion Memory in a closed-box manner with top-one or top-five prediction(s) (\texttt{MM\_Planner\_C\_1} and \texttt{MM\_Planner\_C\_5}), and also assisted in an open-box manner (\texttt{MM\_Planner\_O\_1} and \texttt{MM\_Planner\_O\_5}). 

The results show that  Motion Memory can significantly reduce planning time compared to all three default baseline planners, with the only exception of using Motion Memory in a closed-box manner with Follow in the Trap environment. 
In each of the 12 sub-figures in Fig.~\ref{fig::12_figures}, we can observe increasing performance from left to right. Using Motion Memory in a open-box manner achieves similar or more improvement compared to using it in a closed-box manner. In most cases, using the top-five Motion Memory predictions outperforms using the top-one prediction, despite the potentially more computation to process four extra predictions. 
Comparing the first three columns, Motion Memory can achieve the most significant improvement for Curves environments with GUST and Follow, while RRT enjoys the most Motion Memory benefits in the Trap environments. 
In the last column where Motion Memory does not distinguish among the three classes of planning problems, it also outperforms all three baseline planners. 
Comparing the three rows, Motion Memory is the most helpful for GUST, while the difference made by Motion Memory to RRT is relatively smaller.

\subsection{Continual Improvement with Increasing Experience}

\begin{figure}
    \centering
    
    \begin{minipage}[b]{0.48\columnwidth}
        \centering
        \includegraphics[height=0.7\columnwidth]{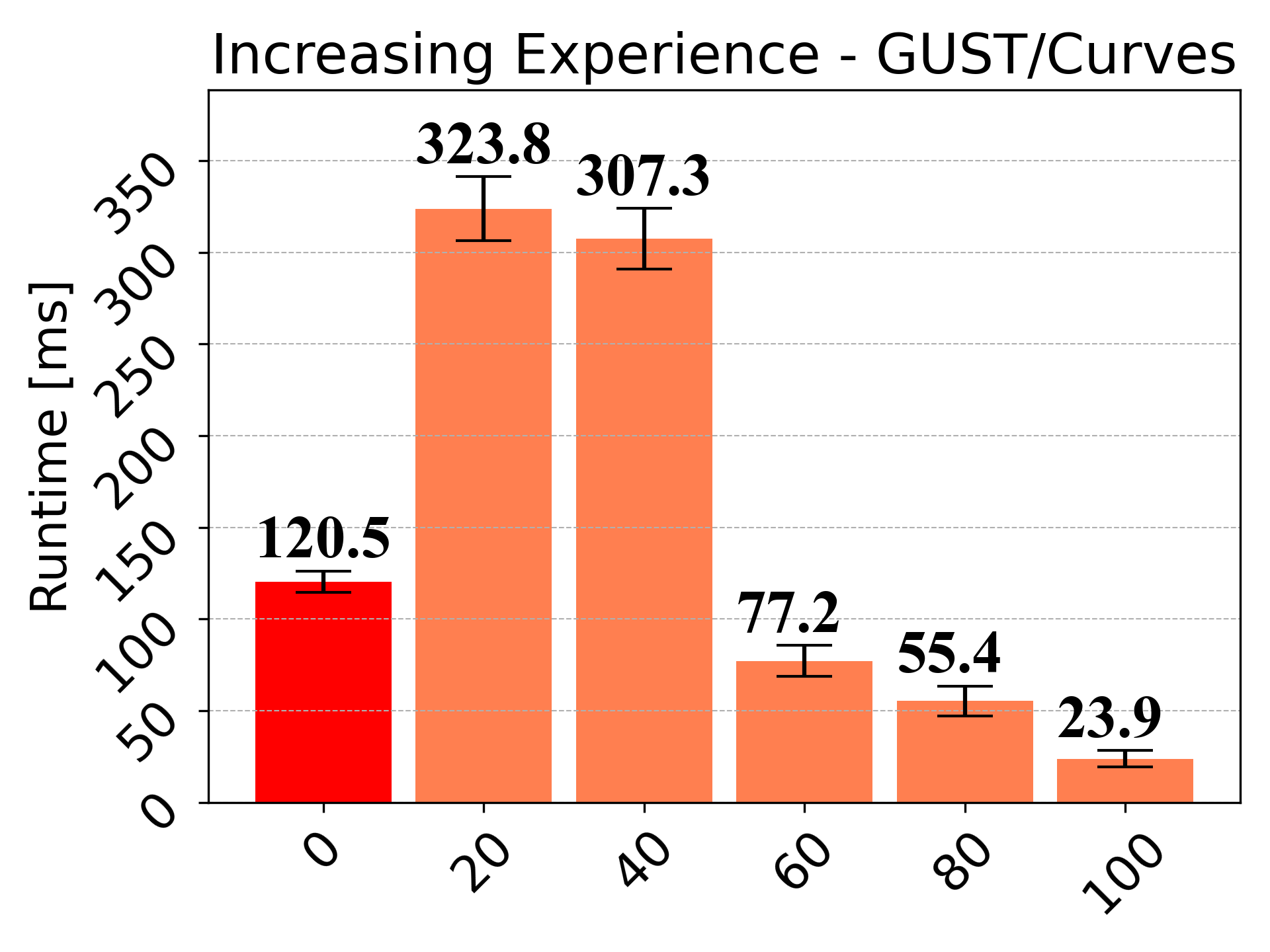}
        \caption{Improvement with Increasing Experiences.}
        \label{fig::incr_exp_fig}
    \end{minipage}
    \hfill
    \begin{minipage}[b]{0.48\columnwidth}
        \centering
        \includegraphics[height=0.7\columnwidth]{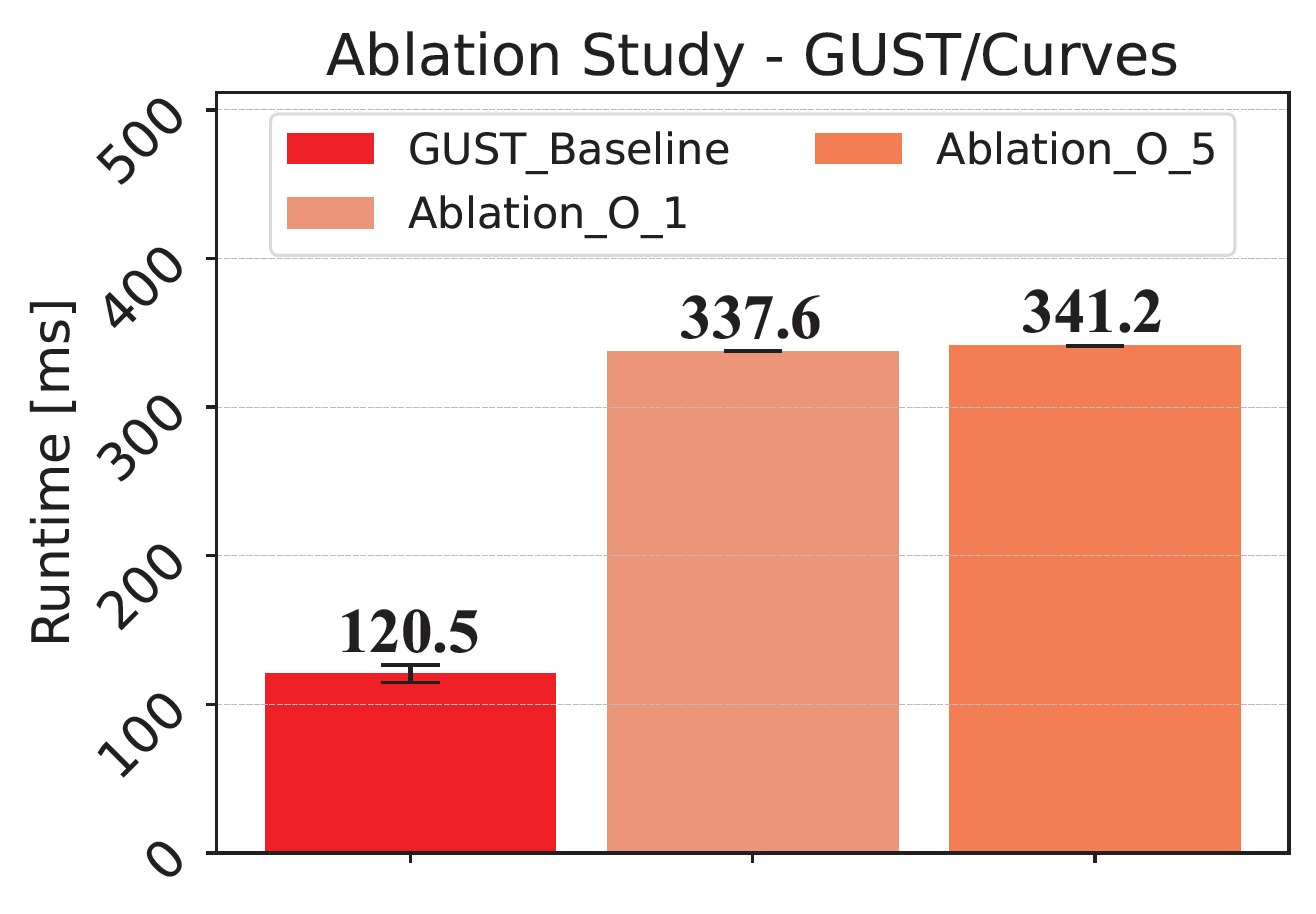}
        \caption{Ablation Study: Random Path Selection.}
        \label{fig::ablation_fig}
    \end{minipage}
    
\end{figure}



We also study the performance improvement of Motion Memory with respect to increasing past planning experiences. In Fig.~\ref{fig::incr_exp_fig}, we show how the runtime of the $\text{GUST}_\text{ClosedBox}$ planner to solve planning problems in Curves environments changes when having access to more available motion plans. For each bar in Fig.~\ref{fig::incr_exp_fig}, we go through the entire Motion Memory pipeline, including environment generation and representation learning, with a limited number of available motion plans (20, 40, 60, 80, or 100). By integrating the Motion Memory model produced by the corresponding amount of past experiences with $\text{GUST}_\text{ClosedBox}$, we test the planner performance on the 10,000 unseen test problems. With only 20 or 40 available past motion plans, Motion Memory underperforms the baseline GUST, because a limited set of available motion plans is not able to sufficiently cover the variety of new planning problems. With increasing experiences, we see a significant reduction in runtime and improvement in planning efficiency. Such experiment results confirm our hypothesis that a motion planner can continually improve when having access to increasing amount of past planning experiences with the assistance of Motion Memory.

\subsection{Ablation Study}



Finally, to demonstrate the necessity of the Motion Memory technique, in addition to the access to a large amount of past planning experiences, we conduct an ablation study, in which we do not use the environment generation and representation learning in Motion Memory to decide which past plan may be most helpful for a future planning problem, but randomly pick a past plan to assist the motion planner. Fig.~\ref{fig::ablation_fig} shows that randomly picking one or five past plans to assist GUST in Curves environments in a open-box manner will increase the planning time. Such results indicate it is necessary to use Motion Memory to decide which past planning experience is useful to accelerate future planning. 
\section{DISCUSSION}
\label{sec::conclusions}

We present Motion Memory, a universal technique which is agnostic to different underlying motion planners but helps them to accelerate their future planning runtime with past planning experiences. By augmenting past experiences and using representation learning, Motion Memory avoids unnecessary and repetitive replanning from scratch when facing similar future planning problems. We demonstrate the efficacy of Motion Memory by integrating it with different planners in a closed-box and open-box fashion, and solving different classes of motion-planning problems more efficiently.  
One possible direction for future research is to extend Motion Memory for manipulation planning, where the robot interacts with the objects in the environment. Another direction is to consider a heterogeneous team of robots and how Motion Memory can facilitate planning for different types of robot, possibly adpating plans from one robot type to another.


\bibliographystyle{IEEEtran}
\bibliography{IEEEabrv,references}

\begin{thebibliography}{10}
\providecommand{\url}[1]{#1}
\csname url@samestyle\endcsname
\providecommand{\newblock}{\relax}
\providecommand{\bibinfo}[2]{#2}
\providecommand{\BIBentrySTDinterwordspacing}{\spaceskip=0pt\relax}
\providecommand{\BIBentryALTinterwordstretchfactor}{4}
\providecommand{\BIBentryALTinterwordspacing}{\spaceskip=\fontdimen2\font plus
\BIBentryALTinterwordstretchfactor\fontdimen3\font minus
  \fontdimen4\font\relax}
\providecommand{\BIBforeignlanguage}[2]{{%
\expandafter\ifx\csname l@#1\endcsname\relax
\typeout{** WARNING: IEEEtran.bst: No hyphenation pattern has been}%
\typeout{** loaded for the language `#1'. Using the pattern for}%
\typeout{** the default language instead.}%
\else
\language=\csname l@#1\endcsname
\fi
#2}}
\providecommand{\BIBdecl}{\relax}
\BIBdecl

\bibitem{canny1988complexity}
J.~Canny, \emph{The complexity of robot motion planning}.\hskip 1em plus 0.5em
  minus 0.4em\relax MIT press, 1988.

\bibitem{kuwata2009real}
Y.~Kuwata, J.~Teo, G.~Fiore, S.~Karaman, E.~Frazzoli, and J.~P. How,
  ``Real-time motion planning with applications to autonomous urban driving,''
  \emph{IEEE Transactions on control systems technology}, vol.~17, no.~5, pp.
  1105--1118, 2009.

\bibitem{fox1997dynamic}
D.~Fox, W.~Burgard, and S.~Thrun, ``The dynamic window approach to collision
  avoidance,'' \emph{IEEE Robotics \& Automation Magazine}, vol.~4, no.~1, pp.
  23--33, 1997.

\bibitem{murray2016robot}
S.~Murray, W.~Floyd-Jones, Y.~Qi, D.~J. Sorin, and G.~D. Konidaris, ``Robot
  motion planning on a chip.'' in \emph{Robotics: Science and Systems}, vol.~6,
  2016.

\bibitem{volpe2003rover}
R.~Volpe, ``Rover functional autonomy development for the mars mobile science
  laboratory,'' in \emph{Proceedings of the 2003 IEEE Aerospace Conference},
  vol.~2, 2003, pp. 643--652.

\bibitem{kavraki1996probabilistic}
L.~E. Kavraki, P.~Svestka, J.-C. Latombe, and M.~H. Overmars, ``Probabilistic
  roadmaps for path planning in high-dimensional configuration spaces,''
  \emph{IEEE transactions on Robotics and Automation}, vol.~12, no.~4, pp.
  566--580, 1996.

\bibitem{hsu1997path}
D.~Hsu, J.-C. Latombe, and R.~Motwani, ``Path planning in expansive
  configuration spaces,'' in \emph{Proceedings of international conference on
  robotics and automation}, vol.~3.\hskip 1em plus 0.5em minus 0.4em\relax
  IEEE, 1997, pp. 2719--2726.

\bibitem{lavalle1998rapidly}
S.~LaValle, ``Rapidly-exploring random trees: A new tool for path planning,''
  \emph{Research Report 9811}, 1998.

\bibitem{karaman2011sampling}
S.~Karaman and E.~Frazzoli, ``Sampling-based algorithms for optimal motion
  planning,'' \emph{The international journal of robotics research}, vol.~30,
  no.~7, pp. 846--894, 2011.

\bibitem{xiao2022motion}
X.~Xiao, B.~Liu, G.~Warnell, and P.~Stone, ``Motion planning and control for
  mobile robot navigation using machine learning: a survey,'' \emph{Autonomous
  Robots}, vol.~46, no.~5, pp. 569--597, 2022.

\bibitem{book:MP}
H.~Choset, K.~M. Lynch, S.~Hutchinson, G.~Kantor, W.~Burgard, L.~E. Kavraki,
  and S.~Thrun, \emph{Principles of Robot Motion: Theory, Algorithms, and
  Implementations}.\hskip 1em plus 0.5em minus 0.4em\relax {MIT} Press, 2005.

\bibitem{book:LaValle}
S.~M. LaValle, \emph{Planning Algorithms}.\hskip 1em plus 0.5em minus
  0.4em\relax Cambridge, MA: Cambridge University Press, 2006.

\bibitem{xiao2021learning}
X.~Xiao, J.~Biswas, and P.~Stone, ``Learning inverse kinodynamics for accurate
  high-speed off-road navigation on unstructured terrain,'' \emph{IEEE Robotics
  and Automation Letters}, vol.~6, no.~3, pp. 6054--6060, 2021.

\bibitem{karnan2022vi}
H.~Karnan, K.~S. Sikand, P.~Atreya, S.~Rabiee, X.~Xiao, G.~Warnell, P.~Stone,
  and J.~Biswas, ``Vi-ikd: High-speed accurate off-road navigation using
  learned visual-inertial inverse kinodynamics,'' in \emph{2022 IEEE/RSJ
  International Conference on Intelligent Robots and Systems (IROS)}.\hskip 1em
  plus 0.5em minus 0.4em\relax IEEE, 2022, pp. 3294--3301.

\bibitem{atreya2022high}
P.~Atreya, H.~Karnan, K.~S. Sikand, X.~Xiao, S.~Rabiee, and J.~Biswas,
  ``High-speed accurate robot control using learned forward kinodynamics and
  non-linear least squares optimization,'' in \emph{2022 IEEE/RSJ International
  Conference on Intelligent Robots and Systems (IROS)}.\hskip 1em plus 0.5em
  minus 0.4em\relax IEEE, 2022, pp. 11\,789--11\,795.

\bibitem{datar2023learning}
A.~Datar, C.~Pan, and X.~Xiao, ``Learning to model and plan for wheeled
  mobility on vertically challenging terrain,'' \emph{arXiv preprint
  arXiv:2306.11611}, 2023.

\bibitem{RRT}
S.~M. LaValle and J.~J. Kuffner, ``Randomized kinodynamic planning,''
  \emph{International Journal of Robotics Research}, vol.~20, no.~5, pp.
  378--400, 2001.

\bibitem{RRTRecent1}
S.~M. LaValle, ``Motion planning: The essentials,'' \emph{IEEE Robotics \&
  Automation Magazine}, vol.~18, no.~1, pp. 79--89, 2011.

\bibitem{RRTguided}
W.~Xinyu, L.~Xiaojuan, G.~Yong, S.~Jiadong, and W.~Rui, ``Bidirectional
  potential guided {RRT}* for motion planning,'' \emph{IEEE Access}, vol.~7,
  pp. 95\,046--95\,057, 2019.

\bibitem{RRTreach}
A.~Shkolnik, M.~Walter, and R.~Tedrake, ``Reachability-guided sampling for
  planning under differential constraints,'' in \emph{IEEE International
  Conference on Robotics and Automation}, 2009, pp. 2859--2865.

\bibitem{RRTtrans2}
D.~Devaurs, T.~Simeon, and J.~Cort{\'e}s, ``Enhancing the transition-based
  {RRT} to deal with complex cost spaces,'' in \emph{IEEE International
  Conference on Robotics and Automation}, 2013, pp. 4120--4125.

\bibitem{EST}
D.~Hsu, J.-C. Latombe, and R.~Motwani, ``Path planning in expansive
  configuration spaces,'' in \emph{IEEE International Conference on Robotics
  and Automation}, vol.~3, 1997, pp. 2719--2726.

\bibitem{KPIECE}
I.~A. \c{S}ucan and L.~E. Kavraki, ``A sampling-based tree planner for systems
  with complex dynamics,'' \emph{IEEE Transactions on Robotics}, vol.~28,
  no.~1, pp. 116--131, 2012.

\bibitem{GUST}
E.~Plaku, ``Region-guided and sampling-based tree search for motion planning
  with dynamics,'' \emph{IEEE Transactions on Robotics}, vol.~31, pp. 723--735,
  2015.

\bibitem{Follow}
E.~Plaku, E.~Plaku, and P.~Simari, ``Clearance-driven motion planning for
  mobile robots with differential constraints,'' \emph{Robotica}, vol.~36, pp.
  971--993, 2018.

\bibitem{mcmahon2022survey}
T.~McMahon, A.~Sivaramakrishnan, E.~Granados, K.~E. Bekris \emph{et~al.}, ``A
  survey on the integration of machine learning with sampling-based motion
  planning,'' \emph{Foundations and Trends{\textregistered} in Robotics},
  vol.~9, no.~4, pp. 266--327, 2022.

\bibitem{aoude2013probabilistically}
G.~S. Aoude, B.~D. Luders, J.~M. Joseph, N.~Roy, and J.~P. How,
  ``Probabilistically safe motion planning to avoid dynamic obstacles with
  uncertain motion patterns,'' \emph{Autonomous Robots}, vol.~35, pp. 51--76,
  2013.

\bibitem{burns2005sampling}
B.~Burns and O.~Brock, ``Sampling-based motion planning using predictive
  models,'' in \emph{Proceedings of the 2005 IEEE international conference on
  robotics and automation}.\hskip 1em plus 0.5em minus 0.4em\relax IEEE, 2005,
  pp. 3120--3125.

\bibitem{kingston2019exploring}
Z.~Kingston, M.~Moll, and L.~E. Kavraki, ``Exploring implicit spaces for
  constrained sampling-based planning,'' \emph{The International Journal of
  Robotics Research}, vol.~38, no. 10-11, pp. 1151--1178, 2019.

\bibitem{sutanto2021learning}
G.~Sutanto, I.~R. Fern{\'a}ndez, P.~Englert, R.~K. Ramachandran, and
  G.~Sukhatme, ``Learning equality constraints for motion planning on
  manifolds,'' in \emph{Conference on Robot Learning}.\hskip 1em plus 0.5em
  minus 0.4em\relax PMLR, 2021, pp. 2292--2305.

\bibitem{baldwin2010non}
I.~Baldwin and P.~Newman, ``Non-parametric learning for natural plan
  generation,'' in \emph{2010 IEEE/RSJ International Conference on Intelligent
  Robots and Systems}.\hskip 1em plus 0.5em minus 0.4em\relax IEEE, 2010, pp.
  4311--4317.

\bibitem{zucker2008adaptive}
M.~Zucker, J.~Kuffner, and J.~A. Bagnell, ``Adaptive workspace biasing for
  sampling-based planners,'' in \emph{2008 IEEE International Conference on
  Robotics and Automation}.\hskip 1em plus 0.5em minus 0.4em\relax IEEE, 2008,
  pp. 3757--3762.

\bibitem{ichter2018learning}
B.~Ichter, J.~Harrison, and M.~Pavone, ``Learning sampling distributions for
  robot motion planning,'' in \emph{2018 IEEE International Conference on
  Robotics and Automation (ICRA)}.\hskip 1em plus 0.5em minus 0.4em\relax IEEE,
  2018, pp. 7087--7094.

\bibitem{hauser2015lazy}
K.~Hauser, ``Lazy collision checking in asymptotically-optimal motion
  planning,'' in \emph{2015 IEEE international conference on robotics and
  automation (ICRA)}.\hskip 1em plus 0.5em minus 0.4em\relax IEEE, 2015, pp.
  2951--2957.

\bibitem{mandalika2019generalized}
A.~Mandalika, S.~Choudhury, O.~Salzman, and S.~Srinivasa, ``Generalized lazy
  search for robot motion planning: Interleaving search and edge evaluation via
  event-based toggles,'' in \emph{Proceedings of the International Conference
  on Automated Planning and Scheduling}, vol.~29, 2019, pp. 745--753.

\bibitem{bialkowski2016efficient}
J.~Bialkowski, M.~Otte, S.~Karaman, and E.~Frazzoli, ``Efficient collision
  checking in sampling-based motion planning via safety certificates,''
  \emph{The International Journal of Robotics Research}, vol.~35, no.~7, pp.
  767--796, 2016.

\bibitem{huh2016learning}
J.~Huh and D.~D. Lee, ``Learning high-dimensional mixture models for fast
  collision detection in rapidly-exploring random trees,'' in \emph{2016 IEEE
  International Conference on Robotics and Automation (ICRA)}.\hskip 1em plus
  0.5em minus 0.4em\relax IEEE, 2016, pp. 63--69.

\bibitem{das2020learning}
N.~Das and M.~Yip, ``Learning-based proxy collision detection for robot motion
  planning applications,'' \emph{IEEE Transactions on Robotics}, vol.~36,
  no.~4, pp. 1096--1114, 2020.

\bibitem{yu2021reducing}
C.~Yu and S.~Gao, ``Reducing collision checking for sampling-based motion
  planning using graph neural networks,'' \emph{Advances in Neural Information
  Processing Systems}, vol.~34, pp. 4274--4289, 2021.

\bibitem{pan2013faster}
J.~Pan, S.~Chitta, and D.~Manocha, ``Faster sample-based motion planning using
  instance-based learning,'' in \emph{Algorithmic Foundations of Robotics X:
  Proceedings of the Tenth Workshop on the Algorithmic Foundations of
  Robotics}.\hskip 1em plus 0.5em minus 0.4em\relax Springer, 2013, pp.
  381--396.

\bibitem{bhardwaj2021leveraging}
M.~Bhardwaj, S.~Choudhury, B.~Boots, and S.~Srinivasa, ``Leveraging experience
  in lazy search,'' \emph{Autonomous Robots}, vol.~45, pp. 979--996, 2021.

\bibitem{hou2020posterior}
B.~Hou, S.~Choudhury, G.~Lee, A.~Mandalika, and S.~S. Srinivasa, ``Posterior
  sampling for anytime motion planning on graphs with expensive-to-evaluate
  edges,'' in \emph{2020 IEEE International Conference on Robotics and
  Automation (ICRA)}.\hskip 1em plus 0.5em minus 0.4em\relax IEEE, 2020, pp.
  4266--4272.

\bibitem{berenson2012robot}
D.~Berenson, P.~Abbeel, and K.~Goldberg, ``A robot path planning framework that
  learns from experience,'' in \emph{2012 IEEE International Conference on
  Robotics and Automation}.\hskip 1em plus 0.5em minus 0.4em\relax IEEE, 2012,
  pp. 3671--3678.

\bibitem{pairet2021path}
{\`E}.~Pairet, C.~Chamzas, Y.~Petillot, and L.~E. Kavraki, ``Path planning for
  manipulation using experience-driven random trees,'' \emph{IEEE Robotics and
  Automation Letters}, vol.~6, no.~2, pp. 3295--3302, 2021.

\bibitem{coleman2015experience}
D.~Coleman, I.~A. {\c{S}}ucan, M.~Moll, K.~Okada, and N.~Correll,
  ``Experience-based planning with sparse roadmap spanners,'' in \emph{2015
  IEEE International Conference on Robotics and Automation (ICRA)}.\hskip 1em
  plus 0.5em minus 0.4em\relax IEEE, 2015, pp. 900--905.

\bibitem{chamzas2019using}
C.~Chamzas, A.~Shrivastava, and L.~E. Kavraki, ``Using local experiences for
  global motion planning,'' in \emph{2019 International Conference on Robotics
  and Automation (ICRA)}.\hskip 1em plus 0.5em minus 0.4em\relax IEEE, 2019,
  pp. 8606--8612.

\bibitem{finney2007predicting}
S.~Finney, L.~P. Kaelbling, and T.~Lozano-P{\'e}rez, ``Predicting partial paths
  from planning problem parameters.'' in \emph{Robotics: Science and
  Systems}.\hskip 1em plus 0.5em minus 0.4em\relax Citeseer, 2007.

\bibitem{qureshi2019motion}
A.~H. Qureshi, A.~Simeonov, M.~J. Bency, and M.~C. Yip, ``Motion planning
  networks,'' in \emph{2019 International Conference on Robotics and Automation
  (ICRA)}.\hskip 1em plus 0.5em minus 0.4em\relax IEEE, 2019, pp. 2118--2124.

\bibitem{johnson2020dynamically}
J.~J. Johnson, L.~Li, F.~Liu, A.~H. Qureshi, and M.~C. Yip, ``Dynamically
  constrained motion planning networks for non-holonomic robots,'' in
  \emph{2020 IEEE/RSJ International Conference on Intelligent Robots and
  Systems (IROS)}.\hskip 1em plus 0.5em minus 0.4em\relax IEEE, 2020, pp.
  6937--6943.

\bibitem{strudel2021learning}
R.~Strudel, R.~G. Pinel, J.~Carpentier, J.-P. Laumond, I.~Laptev, and
  C.~Schmid, ``Learning obstacle representations for neural motion planning,''
  in \emph{Conference on Robot Learning}.\hskip 1em plus 0.5em minus
  0.4em\relax PMLR, 2021, pp. 355--364.

\bibitem{qureshi2020neural}
A.~H. Qureshi, J.~Dong, A.~Choe, and M.~C. Yip, ``Neural manipulation planning
  on constraint manifolds,'' \emph{IEEE Robotics and Automation Letters},
  vol.~5, no.~4, pp. 6089--6096, 2020.

\bibitem{chamzas2021learning}
C.~Chamzas, Z.~Kingston, C.~Quintero-Pe{\~n}a, A.~Shrivastava, and L.~E.
  Kavraki, ``Learning sampling distributions using local 3d workspace
  decompositions for motion planning in high dimensions,'' in \emph{2021 IEEE
  International Conference on Robotics and Automation (ICRA)}.\hskip 1em plus
  0.5em minus 0.4em\relax IEEE, 2021, pp. 1283--1289.

\bibitem{lien2009planning}
J.-M. Lien and Y.~Lu, ``Planning motion in environments with similar
  obstacles.'' in \emph{Robotics: Science and systems}, 2009.

\bibitem{chamzas2022learning}
C.~Chamzas, A.~Cullen, A.~Shrivastava, and L.~E. Kavraki, ``Learning to
  retrieve relevant experiences for motion planning,'' in \emph{2022
  International Conference on Robotics and Automation (ICRA)}.\hskip 1em plus
  0.5em minus 0.4em\relax IEEE, 2022, pp. 7233--7240.

\bibitem{latombe2012robot}
J.-C. Latombe, \emph{Robot motion planning}.\hskip 1em plus 0.5em minus
  0.4em\relax Springer Science \& Business Media, 2012, vol. 124.

\bibitem{xiao2021toward}
X.~Xiao, B.~Liu, G.~Warnell, and P.~Stone, ``Toward agile maneuvers in highly
  constrained spaces: Learning from hallucination,'' \emph{IEEE Robotics and
  Automation Letters}, vol.~6, no.~2, pp. 1503--1510, 2021.

\bibitem{xiao2021agile}
X.~Xiao, B.~Liu, and P.~Stone, ``Agile robot navigation through hallucinated
  learning and sober deployment,'' in \emph{2021 IEEE international conference
  on robotics and automation (ICRA)}.\hskip 1em plus 0.5em minus 0.4em\relax
  IEEE, 2021, pp. 7316--7322.

\bibitem{wang2021agile}
Z.~Wang, X.~Xiao, A.~J. Nettekoven, K.~Umasankar, A.~Singh, S.~Bommakanti,
  U.~Topcu, and P.~Stone, ``From agile ground to aerial navigation: Learning
  from learned hallucination,'' in \emph{2021 IEEE/RSJ International Conference
  on Intelligent Robots and Systems (IROS)}.\hskip 1em plus 0.5em minus
  0.4em\relax IEEE, 2021, pp. 148--153.

\bibitem{lembono2020memory}
T.~S. Lembono, A.~Paolillo, E.~Pignat, and S.~Calinon, ``Memory of motion for
  warm-starting trajectory optimization,'' \emph{IEEE Robotics and Automation
  Letters}, vol.~5, no.~2, pp. 2594--2601, 2020.

\end{thebibliography}
\end{document}